# Data Selection for Semi-Supervised Learning

Shafigh Parsazad[1], Ehsan Saboori[2] and Amin Allahyar[3]

[1] Department Of Computer Engineering, Ferdowsi University of Mashhad
Mashhad, Iran
*Shafigh.Parsazad@stu-mail.um.ac.ir*

[2] K.N Toosi University of Technology of Tehran
Tehran, Iran
*Ehsan.Saboori@ieee.org*

[3] Department Of Computer Engineering, Ferdowsi University of Mashhad
Mashhad, Iran
*amin.allahyar@stu-mail.um.ac.ir*

**Abstract**

The real challenge in pattern recognition task and machine learning process is to train a discriminator using labeled data and use it to distinguish between future data as accurate as possible. However, most of the problems in the real world have numerous data, which labeling them is a cumbersome or even an impossible matter. Semi-supervised learning is one approach to overcome these types of problems. It uses only a small set of labeled with the company of huge remain and unlabeled data to train the discriminator. In semi-supervised learning, it is very essential that which data is labeled and depend on position of data it effectiveness changes. In this paper, we proposed an evolutionary approach called Artificial Immune System (AIS) to determine which data is better to be labeled to get the high quality data. The experimental results represent the effectiveness of this algorithm in finding these data points.

***Keywords:*** *machine learning, semi-supervised learning, evolutionary algorithm, artificial immune system, data selection.*

## 1. Introduction

From the birth of computers the main goal of research was to automation of wide variety of processes that was done with hand before. In general, all automations are appealing because it eliminates the needs of humans to do the task so it reduces time and costs effectively. Computer needs algorithm to do any task. But because of nature of some problems, there is no straight algorithm to solve the problem. For example in process of recognizing a person's face we can detect our relative's picture between hundreds of pictures with no difficulty. But we can't explain how we do this recognition. Another example in this regard is when we can recognize voice of our friend in a very crowded room. Again, there is no explanation in how we do this recognition. Because of this, there is no algorithm to do face recognition task. Instead of using algorithms we can use a special approach called Machine Learning. Machine learning is the process of making a computer system to optimize a performance or solve a problem using example data or experience.

Most of the machine learning tasks consists of training a discriminator with available data which their labels are present and using this discriminator to predict the labels of future and unseen data. The labeled data that is used in training phase is called Train Data and the future data whose label is going to be predicted is called Test Data. The trained discriminator is called Classifier. Another criterion of problem that solves using machine learning approach is when the problem assumption or parameters changes in time or they are depend on a specific environment. Our goal in this regard is to build a training process which can adapt itself with changes in parameters or environment as much as possible. This is called Generalization [1].

Another challenge in machine learning is data acquisition. These data are the hearth of learning process. Because of this, the more high quality training data is the more accurate classifier is trained [2]. Quality in this phrase is consisting of following aspect:

1.1 Noise

In machine learning, the most efficient and important parameter that affect the training accuracy is noise in data. A lot of contribution is made in noise removal in data. This process has many names in researches including: Novelty Detection [3-5], Outlier Detection [6-8] and One Class Classification [9-10]. The most recent and popular methodology to do remove noise from data, is Support Vector Data Description (SVDD) [11]. SVDD effectiveness is measured and proved to be accurate in many case with a high degree of success [12-15].

## 1.2 Missing values

Another very challenging criterion is missing values. Regularly it happens in measurement of data. Because of this, Data measurement instrument play an important role in data acquisition process and quality of these instruments represent quality of result data. But as the accuracy of these instruments raises the price of measuring the data with these instruments also rise, there is a limit on instrument accuracy. Again to overcome this, many scholars tried to find way to predict or approximate these values [16-18]. For example a simple method proposed by Kononenko [19] is to specify the most probable value of missing value in that specific class. Another methodology proposed by Quinlan [20] is using decision tree to predict the missing value of an attribute.

As discussed before the learning task is heavily depend on training data. And the more training data is, the classifier is trained better. So these data should have labels associated with them. But most of the data in real world are raw, unprocessed and unlabeled data. This problem is challenged with a hybrid machine learning method called Semi-Supervised Learning. In this methodology we try to use a very small subset of labeled data and combine them with very large unlabeled data and we try to build a better classifier. As opposed to classification there is Clustering. In clustering we try to split unlabeled data to some groups. Again there is also other approach that is involved in building a clustering system with use a very small subset of labeled data for better accuracy in clustering process. In semi supervised learning the labeled data have a very essential role on training accuracy and it is very depend on which data have label. In this paper we try to determined there data. With this information we can analyze real world data to determine the important data and request our expert to label only these data instead of labeling all of the data in dataset. We used an immune inspired algorithm called aiNet [21] to describe the dataset. After this we labeled these selected data point and fed them to semi supervised learning algorithm. Experimental result showed the effectiveness of this approach in increasing the clustering and classification accuracy using only these selected data.

## 2. Human immune system

The immune system of a human body is a very elaborate and highly accurate system, which is precisely tuned in detecting virus and invaders. Research in this regards shows that we can use this structure in massively-parallel adaptive information-processing system or make our artificial intelligent system more accurate. The worthiness of immune system is because of its specific properties. They are including [22]: Diversity, Distributed, error tolerant, Dynamic, Self-monitoring (self-aware) and Adaptable. With these specifications the immune system will be robust, adaptable and automatic.

1. Robustness: Some system behaviors in a normal environment are the same in the noisy environment. We call these systems Robust or invariant to noise. That is because effect of each piece of immune system is local and independent to others, so even a wrong decision has no big influence in classification accuracy.
2. Adaptable: A system is adaptable if it can change itself with changes in environment and maintain its accuracy as well. Immune system is adaptable because it can detect and recognize new viruses and intruders with a great accuracy.
3. Autonomous: Each immune system component work by itself without any outside governs. This property is a very important feature in the artificial system, because an artificial system should be operational in an unknown environment without the need of constantly guidance of humans.

As discussed, artificial immune system is inspired by human immune system. In this part, we briefly introduce the human immune system [23]. Immune system is consisting of white blood cell which is called lymphocytes. Lymphocytes are divided into two types: B-Cell and T-Cell. These cells are responsible for detection and elimination of any invader or virus which are entering our body. Any subject that activate this system called Antigen or Immunogen. The organs that fight to these antigens are called Antibody. Simply saying, immune system has a memory to save its history of works. With this memory it can perform better next time a same attack happened in the body. Activation of immune system has two phases: primary and secondary:

1. Primary Activation: Primary activation is when the immune system encounters the invader for the first time. In this part large number of antibody will be generated and sent to the detected invader to eliminate the antigen from the body.
2. Secondary Activation: After the invader eliminated from the body, a number of antibodies will still remain in the blood system. These cells are the secondary activation part of the immune system. Because they remember the type of invader and next time these types of invaders entered the body they can operate more effectively. In the other word, they are designed to attack these types of antigens specifically and the body is armed for these antigens. In addition these antibodies will be activated if a similar antigen enters the body. So it is not required to have a specific anybody population for each antigen.

## 3. Ainet algorithm

The aiNet algorithm[23] is inspired by immunity system. In this paper we use a modified version of aiNet algorithm which we called it aiNetDD. We inspired our algorithm from aiNet algorithm proposed by Younsi[24] which we call it *aiNet for Clustering* (aiNetC) for simplicity. He used its algorithm for clustering data. Let $X = \{x_i\}_{i=1}^{N}$ be our dataset, containing $N$ data sample. Each $x_i$ is a vector with $D$ dimension and $x_i \in \mathbb{R}^D$. In aiNet sample data is represented by antigens so we have (1).

$$Ag_i = x_i = \{f_1, f_2, \dots, f_D\} \quad (1)$$

Each B cells is represented with a vector that has same length as antigens or data points (2). They are combined in an antibody network. Let $B = \{b_j\}_{j=1}^{J}$ be our antibody network. With this notation our antibody network is consist of $J$ cells.

$$b_j = \{f_1, f_2, \dots, f_D,\} \quad (2)$$

As discussed before the more the similarity between an antibody and antigen be the strength of connection between these two cell increases. The measure of similarity between antibody and antigen cells is called *Affinity*. We assume an affinity threshold in antibody detection process called *Network Affinity Threshold* (NAT). If the affinity of a given antibody and antigen is lower than NAT, we assume the antibody recognized the antigen. "The aiNetC algorithm is represented below[24]:

1. Generate randomly a population of B cells of size $D$.
2. Read the set of antigens $Ag$ from the data set.
3. Calculate the NAT value using the whole $Ag$ data set.
4. Loop (every member of the set $Ag$):
   4.1. Select an antigen $Ag_i$ from the set $Ag$.
   4.2. For every member of B cell calculate its Euclidean distance to the antigen:
       4.2.1. If the antigen is recognized by a single B cell than store the B cell in memory.
       4.2.2. If the antigen is recognized by more than one B cell then choose the closest and store in memory.
       4.2.3. End For
   4.3. End For
5. Introduce a new population of B cells of size $D$ if no B cell has recognized an antigen and go to step 4.
6. Eliminate that antigen from the sample (hide it from the next generations) Once all the B cells have been presented to that antigen and it has been recognized.
7. Select B cells that have recognized the antigens in this generation.

Mutate B cells and go to step 4 using the mutated cells only (population = mutated B cells)."

We iterate between all antigen until all of them be detected or a number of generations reach. After this phase aiNetC tries to eliminate antibodies that are too close to each other and reduce the memory size until number of B cells becomes equal to number of clusters that is specified. These steps are represented in figure 1.

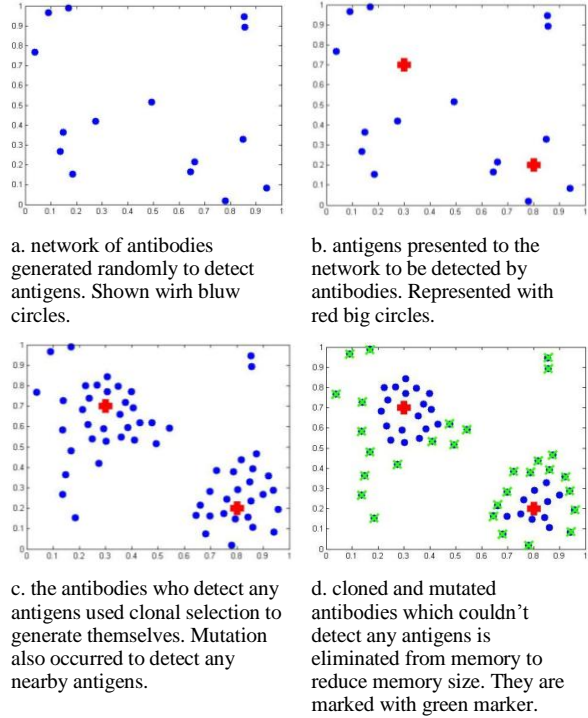

a. network of antibodies generated randomly to detect antigens. Shown wirh bluw circles.

b. antigens presented to the network to be detected by antibodies. Represented with red big circles.

c. the antibodies who detect any antigens used clonal selection to generate themselves. Mutation also occurred to detect any nearby antigens.

d. cloned and mutated antibodies which couldn't detect any antigens is eliminated from memory to reduce memory size. They are marked with green marker.

Fig. 1 aiNet algorithm steps to detect antigens.

## 4. Proposed algorithm

aiNetC algorithm main target, is clustering the given data. We modified the algorithm so that instead of clustering the data it tries to describe them with very few antibodies. The modification done to aiNetC algorithm is shown below:

1. In detection phase 4.2 after an antibody recognized the antigen; we used CLONALG [25] to clone this antibody to reach the more stable memory. This step is like human immune system and is shown in figure 2.
2. A mutation process is added after CLONALG process to detect any nearby antigen that may left.
3. The original antibody is combined with cloned and mutated antibody set. Their affinity again calculated

and the antibody with higher than a pre-defined threshold is selected to be resident in memory. Antibody with low affinity will be eliminated.
4. Very much similar to aiNetC algorithm memory reduction is performed in this phase. All antibodies with an affinity to each other higher than a pre-defined threshold is eliminated.

When the iteration is finished we have a very small set of antibodies that can describe all data point in the dataset.

We argue that if these antibodies have labels in our semi-supervised learning the accuracy of the clustering will be greatly improved. Another advantage of such a method in semi-supervised learning is that labeling the data will not be random and it is done with the lowest possible information that can be provided. In the other word with this approach we get information as lowest as possible while using them as effective as possible.

2.1 Primary Activation

Primary activation is when the immune system encounters the invader for the first time. In this part large number of antibody will be generated and sent to the detected invader to eliminate the antigen from the body.

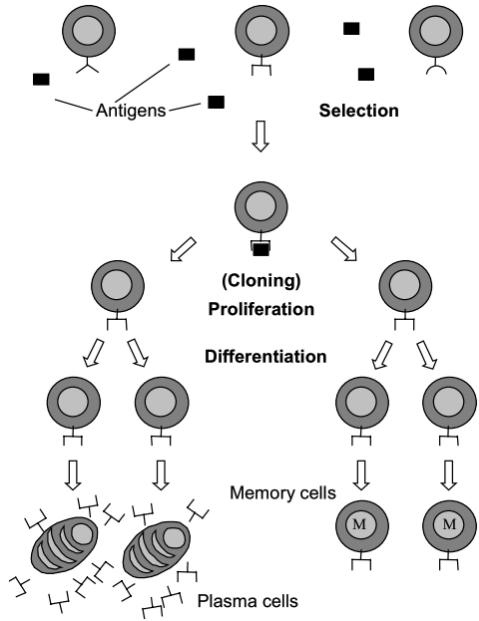

Fig. 2 Demonstration of clonal selection principle.

As debated before, B cell can recognize more than one antigen. These B cells are mutated to recognize more antigen if there is any antigen in the neighborhood. The effectiveness of mutation in finding more local maxima is shown in figure 3. In this figure we can clearly saw that if the mutation is eliminated it is very probable that some local maxima left undetected. The mutation formula used in aiNetC is computed using (3)[26]:

$$m_i = m_i + r \times \Delta \qquad (3)$$

Where $i = \{1, 2, ..., J\}$, $r$ is a random number between $[0,1]$ interval and $\Delta$ is mutation rate pre-defined by user. However aiNetDD is designed for different purpose other than clustering, so this mutation rate should be changed. We used standard mutation rate which is recommended by many researchers including Goldberg and Holland[27]. They argue that for a genome with length of $l$ we have mutation rate equal to $\frac{1}{l}$. Simply saying we want to have at least a mutation in every genome. For calculating mutation value we need it to be very depending on data. For this reason we used dataset range to calculate mutation value. The data range is calculated using (4):

$$Range.Max_{d \in D} = \max\{x_{id}\}_{i=1}^{N}$$
$$Range.Min_{d \in D} = \min\{x_{id}\}_{i=1}^{N} \qquad (4)$$

Where $Range.Max_d$ and $Range.Min_d$ is maximum and minimum range of all data based on dimension $d$.

After evaluation we find that the best result will be when the mutation value is $\frac{1}{20}$ of data range. In the other word we want our antibody can traverse across all data with 20 steps, if it mutate only in one direction. So the mutation value is calculated using (5):

$$mutation\ rate_d = \frac{(Range.Max_d - Range.Min_d)}{20} \qquad (5)$$

In figure 4 result of using aiNetC on a sample dataset and the iris dataset is represented. As shown in this figure all the antibodies if propagated around dataset and we can describe all dataset with only a few point of antibodies. It should be noted that not all antibodies is located in a good spot. To eliminate the redundant antibodies we measured the affinity of each antibody with all antigens in neighborhood of given antibody.

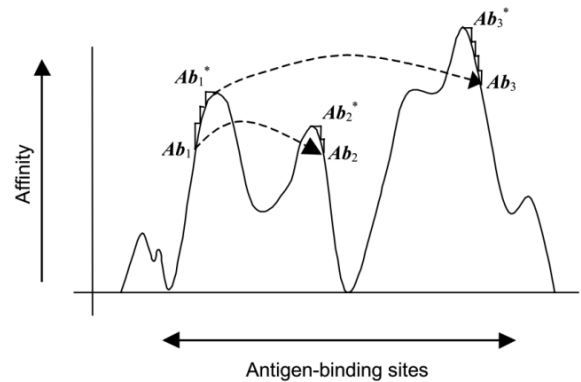

Fig. 3 Representation of how mutation can help on finding nearby antigens.

After this, antibody that has no antigen in its neighborhood is eliminated. In figure 4 antibodies that become memories is illustrated with squares and eliminated antibodies is shown with star with green color.

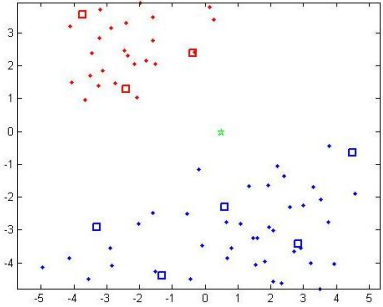

Fig. 4 Result of aiNetC for an artificial two dimensional dataset.

Figure 5 is result of aiNetC algorithm on iris dataset. This dataset is represented with four figures so we can see all dimension of this dataset. Again antibodies are propagated in input space to describe this dataset with high quality. This dataset has three clusters, so that with used three colors to better demonstration of clusters. Same notation as figure 4 is used for memory cells and eliminated antibodies. It should be noted that in evaluation phase label of data is hidden to algorithm. After the algorithm computed the memory cells we used original labels to colorize the figure for better illustration of result.

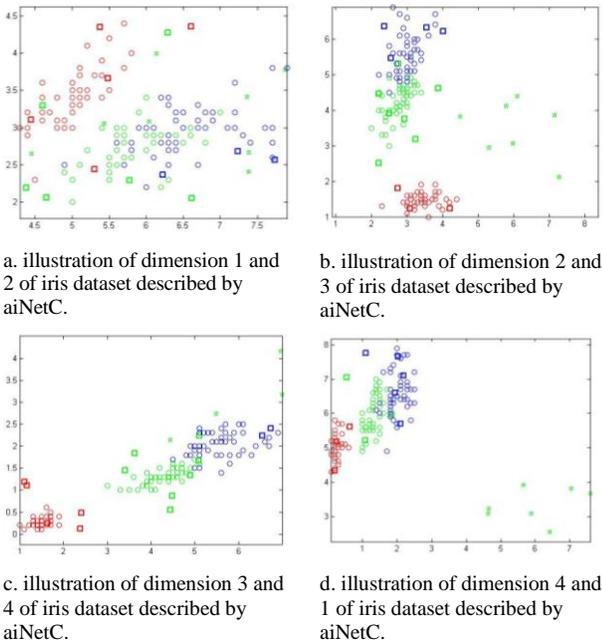

a. illustration of dimension 1 and 2 of iris dataset described by aiNetC.

b. illustration of dimension 2 and 3 of iris dataset described by aiNetC.

c. illustration of dimension 3 and 4 of iris dataset described by aiNetC.

d. illustration of dimension 4 and 1 of iris dataset described by aiNetC.

Fig. 5 Evaluation result of using aiNetC algorithm on iris dataset.

## 4. Experimental result

For experimental result we used two semi-supervised learning algorithm including: semi-supervised KMeans [28] as our clustering algorithm and semi-supervised support vector machines [29] as our classification algorithm. For dataset we used five dataset including: Iris, Soybean, Wine, Digits-389, and Letter-IJL. First of all random set of labels is generated and fed into these algorithms. Different number of labels for each clustering and classification algorithm is used to build a confident result. For classification algorithm we used 10-fold cross validation approach to show the generalization of proposed algorithm. To measure the quality of clustering, accuracy measurement is used, so we have a unique way of comparing algorithms.

After this phase, our algorithm is used for analyze of data and all datasets is fed to aiNetC algorithm. After analyzing aiNetC algorithm recommend some data to be labeled by user to achieve the best result. These labeled data is again fed to the learning algorithms as information that we have from dataset. Result of this learning is measured. A complete representation of outcome of these measurements is shown in below tables. Experimental results denote the effectiveness of such methodology. Although analyzing the dataset before the learning process begins have computational cost but it should be kept in mind that this process is considered as a pre-processing step. In the other word it is an offline process which improvement of data in compare of computational cost seems reasonable.

Table 1: Result of semi supervised classification and semi supervised clustering algorithm with random labels provided for Semi-Supervised KMeans algorithm, represented with accuracy percent.

| Number of Labels | 10 | 20 | 30 | 40 |
|---|---|---|---|---|
| Iris | 91.02 | 92.54 | 92.80 | 93.46 |
| Soybean | 83.51 | 84.10 | 85.19 | 87.63 |
| Wine | 73.30 | 74.24 | 75.15 | 76.84 |
| Digits-389 | **94.17** | **94.25** | **94.30** | **95.74** |
| Letter-IJL | 83.74 | 84.81 | 85.28 | 86.36 |

Table 2: Result of semi supervised classification and semi supervised clustering algorithm with random labels provided for Semi-Supervised Support Vector Machines algorithm, represented with accuracy percent.

| Number of Labels | 10 | 20 | 30 | 40 |
|---|---|---|---|---|
| Iris | **92.96** | 93.52 | 94.86 | 94.96 |
| Soybean | **81.27** | 83.37 | 84.50 | 86.25 |
| Wine | **80.82** | **81.37** | **81.70** | **82.24** |
| Digits-389 | 93.36 | 94.50 | 95.70 | 96.79 |
| Letter-IJL | 81.26 | 82.63 | 82.22 | 84.15 |

Table 3: Result of semi supervised classification and semi supervised clustering algorithm with recommended labels by aiNetC provided for Semi-Supervised KMeans algorithm, represented with accuracy percent.

| Number of Labels | 10 | 20 | 30 | 40 |
|---|---|---|---|---|

| Iris | **92.63** | **93.26** | **94.25** | **95.29** |
|---|---|---|---|---|
| Soybean | 80.25 | 81.28 | 81.60 | 82.64 |
| Wine | 71.26 | 71.62 | 72.17 | 74.88 |
| Digits-389 | 80.92 | 81.03 | 82.83 | 84.17 |
| Letter-IJL | **84.15** | **84.91** | **85.56** | **85.90** |

Table 4: Result of semi supervised classification and semi supervised clustering algorithm with recommended labels by aiNetC provided for Semi-Supervised Support Vector Machines algorithm, represented with accuracy percent.

| Number of Labels | 10 | 20 | 30 | 40 |
|---|---|---|---|---|
| Iris | 92.47 | **94.17** | 94.73 | **94.98** |
| Soybean | 80.97 | **87.68** | **91.82** | **94.03** |
| Wine | 72.73 | 72.97 | 73.19 | 75.55 |
| Digits-389 | 85.17 | 86.39 | 86.76 | 88.92 |
| Letter-IJL | **85.28** | **85.83** | **86.18** | **86.40** |

**Shafigh Parsazad** is graduated from Ferdowsi University of Mashhad, Iran in computer engineering and is graduated from Ferdowsi University of Mashhad, Mashhad, Iran in Artificial Intelligence. He is interested in "Pattern Recognition", "Image Processing", "Network Security" and "Bioinformatics". He is currently working on Nature-inspired algorithms and Computational Biology.

**Ehsan Saboori** is graduated from Ferdowsi University of Mashhad, Iran in computer engineering and is graduated from K.N Toosi University of technology, Tehran, Iran in IT engineering. He is interested in "Peer-to-Peer Networks", "Computer Networks", "Network Security" and "Anonymity". He currently works on peer to peer network security and privacy.